\documentclass[10pt,twocolumn,letterpaper]{article}

\usepackage{cvpr}
\usepackage{times}
\usepackage{epsfig}
\usepackage{graphicx}
\usepackage{amsmath}
\usepackage{amssymb}
\usepackage{booktabs} 
\usepackage{mathrsfs}
\usepackage{multirow,array}
\usepackage{float}
\usepackage{enumitem}
\usepackage{color}

\usepackage[ruled]{algorithm2e} 

\SetAlFnt{\small}
\SetAlCapFnt{\small}
\SetAlCapNameFnt{\small}
\SetAlCapHSkip{0pt}
\IncMargin{-\parindent}

\newcommand{\norm}[2]{\left\lVert#1\right\rVert_{#2}}

\usepackage[pagebackref=true,breaklinks=true,letterpaper=true,colorlinks,bookmarks=false]{hyperref}

\cvprfinalcopy 

\def\cvprPaperID{5863} 

\ifcvprfinal\fi
\begin{document}

\title{Learning to Learn Image Classifiers with Visual Analogy}

\author{Linjun Zhou$^1$ \qquad Peng Cui$^1$ \qquad Shiqiang Yang$^1$ \qquad Wenwu Zhu$^1$ \qquad Qi Tian$^2$\\
$^1$Tsinghua University \qquad $^2$Huawei Noah's Ark Lab\\
\tt\small zhoulj16@mails.tsinghua.edu.cn \\ \tt\small\{cuip, yangshq, wwzhu\}@mail.tsinghua.edu.cn, \quad tian.qi1@huawei.com}

\maketitle
\thispagestyle{empty}

\begin{abstract}
Humans are far better learners who can learn a new concept very fast with only a few samples compared with machines. The plausible mystery making the difference is two fundamental learning mechanisms: learning to learn and learning by analogy. In this paper, we attempt to investigate a new human-like learning method by organically combining these two mechanisms. In particular, we study how to generalize the classification parameters from previously learned concepts to a new concept. we first propose a novel Visual Analogy Graph Embedded Regression (VAGER) model to jointly learn a low-dimensional embedding space and a linear mapping function from the embedding space to classification parameters for base classes. We then propose an out-of-sample embedding method to learn the embedding of a new class represented by a few samples through its visual analogy with base classes and derive the classification parameters for the new class. We conduct extensive experiments on ImageNet dataset and the results show that our method could consistently and significantly outperform state-of-the-art baselines. 
\end{abstract}
\section{Introduction}

The emergence of deep learning has advanced the image classification performance into an unprecedented level. The error rate on ImageNet has been halved and halved again \cite{krizhevsky2012imagenet, simonyan2014very, he2016deep}, even approaching human-level performance. Despite the success, the state-of-the-art models are notoriously data hungry, requiring tons of samples for parameter learning. In real cases, however, the visual phenomena follows a long-tail distribution \cite{Zhu_2014_CVPR} where only a few sub-categories are data-rich and the rest are with limited training samples. How to learn a classifier from as few samples as possible is critical for real applications and fundamental for exploring new learning mechanisms.

Compared with machines, people are far better learners as they are capable of learning models from very limited samples of a new category and make accurate prediction and judgment accordingly. An intuitive example is that a baby learner can learn to recognize a wolf with only a few sample images provided that he/she has been able to successfully recognize a dog. The key mystery making the difference is that people have strong prior knowledge to generalize across different categories \cite{lake2017building}. It means that people do not need to learn a new classifier (\textit{e.g.} wolf) from scratch as most machine learning methods, but generalize and adapt the previously learned classifiers (\textit{e.g.} dog) towards the new category. A major way to acquire the prior knowledge is through learning to learn from previous experience. In the image classification scenario, learning to learn refers to the mechanism that learning to recognize a new concept can be accelerated by previously learned other related concepts.

A typical image classifier is constituted by representation and classification steps, leading to two fundamental problems in learning to learn image classifiers: (1) how to generalize the representations from previous concepts to a new concept, and (2) how to generalize the classification parameters of previous concepts to a new concept. In literature, transfer learning and domain adaptation methods \cite{Patricia_2014_CVPR} are proposed with a similar notion, mainly focusing on the problem of representation generalization across different domains and tasks. With the development of CNN-based image classification models, the high-level representations learned from very large scale labeled dataset are demonstrated to have good transferability across different concepts or even different datasets \cite{tzeng2015simultaneous}, which significantly alleviate the representation generalization problem. However, how to generalize the classification parameters in deep models (\textit{e.g.} the fc7 layer in AlexNet) from well-trained concepts to a new concept (with only a few samples) is largely ignored by previous studies.

Learning by analogy has been proved to be a fundamental building block in human learning process \cite{gentner1997reasoning}, a plausible explanation on the fast learning of novel class is that a human learner selects some similar classes from the base classes by visual analogy, transfers and combines their classification parameters for the novel class. In this sense, visual analogy provides an effective and informative clue for generalizing image classifiers in a way of human-like learning. But the limited number of samples in the new class would cause inaccurate and unstable measurements on visual analogy in high-dimensional representation space, and how to transfer the classification parameters from selected base classes to a new class is also highly non-trivial for the generation efficacy.

\begin{figure*}
\centering
\includegraphics[width=6.0in]{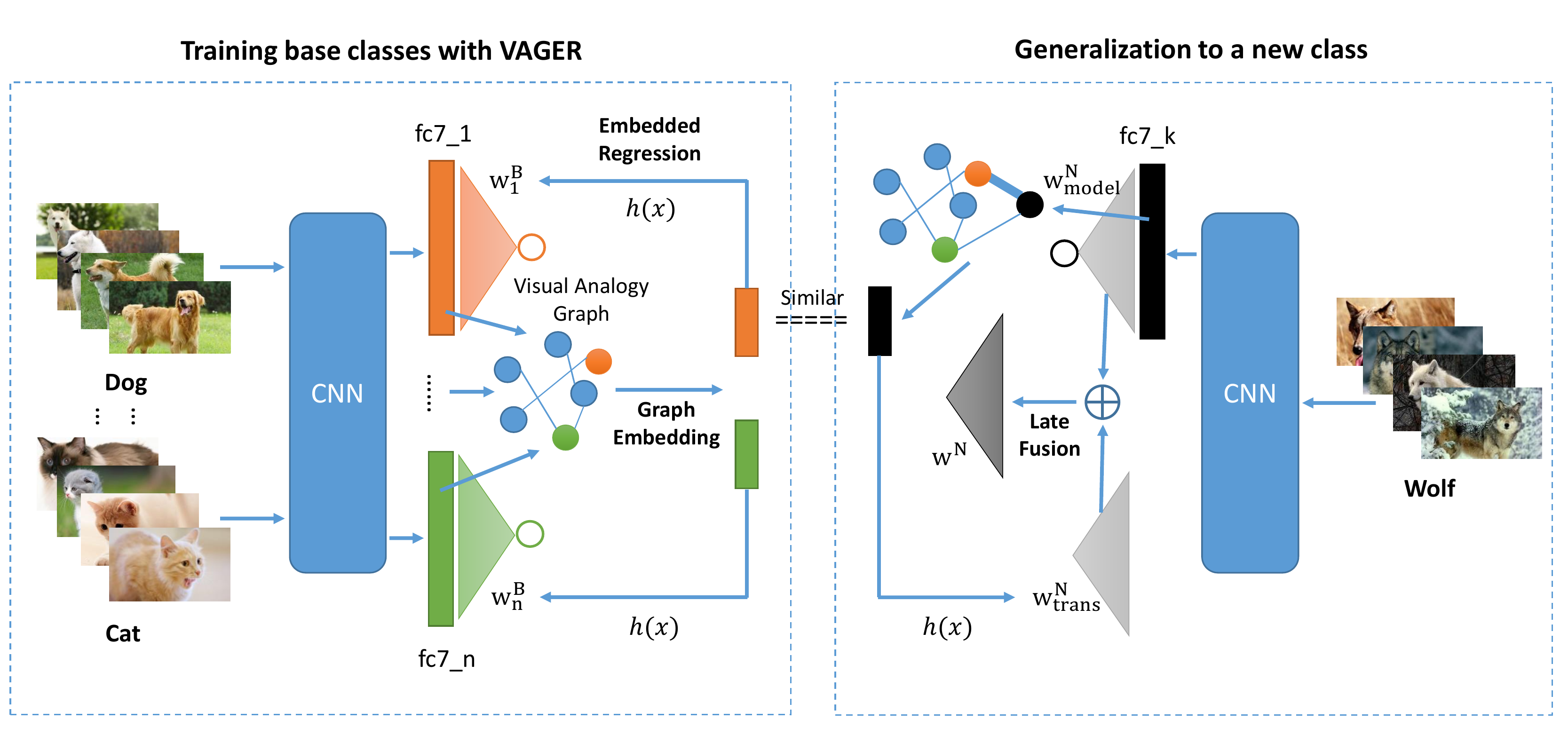}
\caption{The framework of learning to learn image classifiers. \emph{Training Base Classes with VAGER}:  By training base classes with VAGER, we derive the embeddings of each base class and the common mapping function from embeddings to classification parameters. \emph{Generalization to a New Class}: Given a new class with only a few samples, we can infer its embedding through out-of-sample inference, and then transform the embedding into transferred classification parameters by the mapping function learned by VAGER. After training the classifier with new class samples and getting the model classification parameters, we fuse the two kinds of parameters to form the final classifier.}
\label{Fig:Process}
\end{figure*}

To address the above problems, we first propose a novel Visual Analogy Graph Embedded Regression (VAGER) model to jointly learn a low-dimension embedding space and a linear mapping function from the embedding space to classification parameters for base classes. In particular, we learn a low-dimension embedding for each base class so that embedding similarity between two base classes can reflect their visual analogy in the original representation space. Meanwhile, we learn a linear mapping function from the embedding of a base class to its previously learned classification parameters (\textit{i.e.} the logistic regression parameters). The VAGER model enables the transformation from the original representation space to embedding space and further into classification parameters. We then propose an out-of-sample embedding method to learn the embedding of a new class represented by a few samples through its visual analogy with base classes. By inputting the learned embedding into VAGER, we can derive the classification parameters for the new class. Note that these classification parameters are purely generated from base classes (\textit{i.e.} transferred classification parameters), while the samples in the new class, although only a few, can also be exploited to generate a set of classification parameters (\textit{i.e.} model classification parameters). Therefore, we further investigate the fusion strategy of the two kinds of parameters so that the prior knowledge and data knowledge can be fully leveraged. The framework of the proposed method is illustrated in Figure \ref{Fig:Process}.

The technical contributions of this paper are three folds. (1) We introduce the mechanism of visual analogy into image classification, which provides a new way of transferring classification parameters from previous concepts to a new concept.
(2) We propose a novel VAGER model to realize the transformation from original representation to classification parameters for any new class. (3) We intensively evaluate the proposed method and the results show that our method consistently and significantly outperform other baselines.

\section{Related Work}
\noindent
\textbf{One/Few-shot Learning.} \hspace{0.1cm} One/Few-shot learning mainly focuses on how to train models from just one, or a handful of images instead of the large-scale training dataset. \cite{fei2006one} first proposed this concept as well as a transfer method via a Bayesian approach on the low-level visual features. Afterward researchers have been working on hand-crafted visual features. \cite{yao2010boosting, qi2011towards} propose transfer mechanism based on Adaboost-SVM method. They both construct a set of weak classifiers through the data from the base classes and learn a new classifier by linearly combining the weak classifiers. Furthermore,  \cite{tommasi2014learning} proposes an adaptive Least-Square SVM method. These methods require huge supervised information to learn the weight of the combined model and the insufficient representative ability of low-level features limits their performance.

After deep learning is introduced to the large-scale image classification, benefited from its strong representative ability, the performance of the few-shot learning is improved gradually. \cite{koch2015siamese} introduces a two-way Siamese Neural Network to learn the similarity of two input images as the evaluation metric, which is an early work of few-shot learning combined with deep learning.  Afterwards, meta-learning provides a new training mechanism and shows great performance on small datasets like Omniglot \cite{Lake2013One} and MiniImageNet \cite{vinyals2016matching}. MANN\cite{santoro2016one}, Matching Network\cite{vinyals2016matching}, MAML\cite{Finn2017Model}, Prototypical Network\cite{Snell2017Prototypical}, Relation Network\cite{sung2018learning} are some representitive works. Their methods introduce a new training mechanism to completely simulate evaluation circumstance on $m$-way $k$-shot classification, where training data is split into support sets and training process is based on the support set, not a single image. However, they perform not so well on large-scale datasets like ImageNet. For large-scale datasets, \cite{hariharan2017low} proposes a Squared Gradient Magnitude Loss considering both the multi-class logistic loss and small dataset training loss, \cite{wang2016learning} proposes a Model Regression Network for intra-class transfer which learns a nonlinear mapping from the model parameter trained by small-samples to the model parameter trained by large-samples. More recently, a few works exploit generative models to create more data for training. \cite{rezende2016one} takes advantage of the deep generative models to give a method to produce similar images from given images. \cite{WangCVPR2018a} adds a deep hallucinator structure to the original meta-learning methods and trains the hallucinator and the classifier at the same time. 

\noindent
\textbf{Learning to Learn Image Classifiers.} \hspace{0.1cm} {The problem focuses on how to learn classifier parameters for a novel class and the methods are widely used in zero-shot learning and few-shot learning. \cite{Elhoseiny2014Write} and \cite{Ba2015Predicting} use purely textual description of categories to learn the parameter of the classifier in zero-shot image classification. \cite{Elhoseiny2014Write} uses a kernel method to learn from the textual feature to the parameter, while \cite{Ba2015Predicting} uses a neural network. Further, \cite{Changpinyo2016Synthesized} learns base classifiers and construct classifiers of novel classes utilizing attribute similarities between classes. Recently, \cite{qiao2018few} and \cite{Qi2018Low} investigate how to utilize visual features to generate classifier parameters for novel class and show good performance on few-shot learning. Different from these previous works, our work concentrates more on how to generate classification parameters with visual analogy at category level.} 

\noindent
\textbf{Graph Embedding.} \hspace{0.1cm} Graph Embedding (Network Embedding) is used to extract the formalized representation of each node in a large-scale graph or network. The low-dimension hidden embeddings could capture both the relationship between nodes and the features of each node itself. Graph Embedding is widely used in the social network area to solve the node clustering or link prediction problems \textit{etc.} There are many classical algorithms in graph embedding; we list some of them but not all. For example, \cite{ahmed2013distributed} uses a matrix factorization technique which is optimized by SGD and \cite{tang2015line} proposes LINE method which preserves both the first-order and second-order proximities of each node and improves the quality of the embeddings \textit{etc.} Graph embedding is proved to be an effective method in the graph analysis area.

\section{Methodology}

\subsection{Notations and Problem Formulation}
\label{Algorithm:Notations}
Suppose that we have an image set $I$, and the set is divided into base-class set $I^B = I^B_1 \cup I^B_2 \cup \cdots \cup I^B_n$ which have sufficient training samples, and novel-class set $I^N = I^N_1 \cup I^N_2 \cup \cdots \cup I^N_m$ which have only a few training samples in each class. We train an AlexNet \cite{krizhevsky2012imagenet} on $I^B$ as our base CNN model and extract its fc7 layer as the high-level features of images. The feature space is denoted as $\mathscr{X} \subset \mathbb{R}^d$. For each image in $I^B$, we obtain its fc7 layer feature $\mathbf{x}_{ij}^B \in \mathscr{X}$ where $i = 1,2, \cdots , n$ represents its class and $j = 1,2, \cdots, |I^B_i|$ represents its index in class $i$. We use the same CNN model to derive high-level representations for images in novel classes, denoted by {$\mathbf{x}_{ij}^N$.

A typical binary classifier can be represented as $f(\, \cdot \,; \mathbf{w}|\mathbf{X})$ which is a mapping function $f:\mathbb{R}^d \xrightarrow{} \mathbb{R}$ parametrized by $\mathbf{w}$. The input is a $d$-dimensional image feature vector and the output is the probability that the image belongs to the class.  We use $\mathbf{w}_i^B$ to denote the parameters for base class $i$ and $\mathbf{w}_i^N$ for novel class $i$. Based on the above notations, Our problem is defined as follows. 
\newtheorem{problem}{Problem}
\begin{problem}[Learning to learn image classifiers]
\textbf{Given} the image features of base classes $\mathbf{X}^B$, the well-trained base classifier parameters $\mathbf{W}^{B}$, and the image features of a novel class $i$ $\mathbf{X}_i^N$ with only a few positive samples, \textbf{learn} the classification parameters $\mathbf{w}^{N}_i$ for the novel class, so that the learned classifier $f(\, \cdot \,; \mathbf{w}^{N}_i|\mathbf{X}^B, \mathbf{W}^{B}, \mathbf{X}_i^N)$ can precisely predict labels for the $i^{th}$ novel class.
\end{problem}
Note that the problem of learning to learn image classifiers differs from traditional image classification problems in that the learning of a classifier for a novel class depends on the previously learned base-class classifiers and the image representations in base classes besides the image samples in the novel class. 

\subsection{The VAGER Model}
\label{Algorithm:Learning}
We define a graph $G = (V, E)$ where $V$ is the vertex set of the graph, with each vertex representing a base class and $|V| = n$. $E$ is the edge set of the graph, each edge represents visual analogy relationship between two classes with the edge weight depicting the similarity degree. We use $\mathbf{A}$ to represent the adjacency matrix of the network, and $\mathbf{A}_{ij}$ is the edge weight between vertex $i$ and vertex $j$. $\mathbf{A}_{i,:}$ and $\mathbf{A}_{:,j}$ stands for the i-th row and the j-th column of $\mathbf{A}$ respectively. In our classification problem, we construct the visual analogy network as a undirected complete graph, and edge weight (i.e. degree of visual analogy) between two classes is calculated by:
\begin{equation}
\label{equ:sim}
\mathbf{A}_{ij} = \frac{\overline{\mathbf{x}_i^B} \cdot \overline{\mathbf{x}_j^B}}{\Vert \overline{\mathbf{x}_i^B} \Vert_2 \cdot \Vert \overline{\mathbf{x}_j^B} \Vert_2}.
\end{equation}
Here $\overline{\mathbf{x}_i^B}$ means the average feature vector for class $i$ and this equation is the cosine similarity between two base classes. Note that our graph is an undirected graph, and the adjacency matrix $\mathbf{A}$ is symmetric.

To make the visual analogy measurement robust in sparse scenarios, we need to reduce the representation space dimensions. Our basic hypothesis in generalizing classification parameters is that if two classes are visually similar, they should share similar classification parameters. By imposing a linear mapping function from the embedding space to classification parameter space, similar embeddings will result in similar classification parameters. Motivated by this, we propose a Visual Analogy Graph Embedded Regression model. 

Let $\mathbf{V} \in \mathbb{R}^{n \times q}$ be the embeddings for all nodes in the graph, and each row of $\mathbf{V}$ with dimension $q$ is the embedding for each vertex. Let $\mathbf{W} \in \mathbb{R}^{n \times p}$ represent all parameters of the base classifiers. There is also a common linear transformation matrix for all base classes $\mathbf{T} \in \mathbb{R}^{q \times p}$ to convert the embedding space to the classification parameter space for all base classifiers. Then the loss function is defined as:
\begin{equation}
\label{eqn:base}
\mathscr{L}(\mathbf{V}, \mathbf{T}) =  \Vert \mathbf{VT} - \mathbf{W} \Vert_F^2 + \beta \Vert \mathbf{A} - \mathbf{V}\mathbf{V}^\top \Vert_F^2.
\end{equation}
where $\Vert \cdot \Vert_F$ is the Frobenius Norm of the matrix.

The first term enforces the embeddings to be able to convert into the classification parameter through a linear transformation. The second term constrains the embeddings to preserve the structure of the visual analogy graph. Our goal is to find the matrix $\mathbf{V}$ and $\mathbf{T}$ to minimize this loss function.

This is a common unconstrained two variables optimization problem and we use the alternative coordinate descent method to find the best solution for $\mathbf{V}$ and $\mathbf{T}$, where the gradients are calculated by:
\begin{equation}
\left\{
\begin{aligned}
&\frac{\partial \mathscr{L}(\mathbf{V}, \mathbf{T})}{\partial \mathbf{V}} = 2(\mathbf{VT}-\mathbf{W})\mathbf{T}^\top + \beta(-4\mathbf{AV} + 4\mathbf{V}\mathbf{V}^\top \mathbf{V})\\
&\frac{\partial \mathscr{L}(\mathbf{V}, \mathbf{T})}{\partial \mathbf{T}} = 2\mathbf{V}^ \top(\mathbf{VT}-\mathbf{W}).
\end{aligned}
\right.
\end{equation}

\subsection{Embedding Inference for Novel Classes}
\label{Algorithm:Predicting}
By training VAGER model in base classes, we can obtain the embeddings for each base class and the mapping function from embeddings to classification parameters. Given a new class with only a few samples, we need to infer its embedding. Suppose the embedding for the novel class is $\mathbf{v}_{new} \in \mathbb{R}^q$. We calculate the similarity of a novel class with all base classes by Equation \ref{equ:sim}, and we denote this similarity vector by $\mathbf{a}_{new} \in \mathbb{R}^n$.

Then we define the objective function for the novel class embedding inference and our goal is to minimize the following function:
\begin{equation}
\label{eqn:novel}
\mathscr{L}(\mathbf{v}_{new}) = \norm{\begin{bmatrix} \mathbf{A} & \mathbf{a}_{new}^\top \\ \mathbf{a}_{new} & 1 \end{bmatrix} - \begin{bmatrix} \mathbf{V} \\ \mathbf{v}_{new} \end{bmatrix} \begin{bmatrix} \mathbf{V}^\top & \mathbf{v}_{new}^\top \end{bmatrix}}{F}^2.
\end{equation}
Equation \ref{eqn:novel} is in fact the extension of the second term in Equation \ref{eqn:base}. As we have little information about the classification parameters of the novel class, we omit the first term in Equation \ref{eqn:base}.

After we delete the independence term of $\mathbf{v}_{new}$, the final minimization problem for us to solve is:
\begin{equation}
\label{eqn:fixednovel}
\min \mathscr{L}(\mathbf{v}_{new}) = 2\norm{\mathbf{a}_{new} - \mathbf{v}_{new}\mathbf{V}^\top}{2}^2 + (\mathbf{v}_{new}\mathbf{v}_{new}^\top - 1).
\end{equation}
In fact, the second term of Equation \ref{eqn:fixednovel} is a regularization term. We omit the second term and thus the first term is in the form of a linear regression loss. Then we can get the explicit solution for $\mathbf{v}_{new}$ without using gradient descent. The solution is represented as:
\begin{equation}
\label{eqn:produce}
\mathbf{v}_{new} = \mathbf{a}_{new}(\mathbf{V}^\top)^+,
\end{equation}
where $\mathbf{M}^+$ is the Moore-Penrose pseudo-inverse of matrix $\mathbf{M}$ defined by $(\mathbf{M}^\top \mathbf{M})^{-1}\mathbf{M}^\top$. Note that we could speed up the algorithm by pre-computing the pseudo-inverse of $\mathbf{V}^\top$.

After deriving the embedding for the new class, we can easily obtain its transferred classification parameters by multiplying transformation matrix $\mathbf{T}$:
\begin{equation}
\label{eqn:final}
\mathbf{w}_{new}^N = \mathbf{v}_{new} \mathbf{T}.
\end{equation}

\subsection{Parameter Refinement}
\label{Algorithm:Latefusion}
As mentioned above, we can also learn the classification parameters of a new class from its samples (although only a few), and we call them model classification parameters. Then we need to fuse the transferred classification parameters and model classification parameters into the final classifier. Here we present three strategies for refinement: Initializing, Tuning, and Voting. 

Let $f(\cdot, \mathbf{w}^N): \mathbb{R}^d \xrightarrow{} [0,1]$ be the binary classifier for a new class. $\mathbf{X}_{T}$ is the mixture set of positive and negative samples, and $y$ is the label with $y=1$ indicating positive sample and $y=0$ indicating negative sample. 
\\

\noindent
\textbf{Initializing} \ We use the transferred classification parameters as an initialization and then re-learn the parameters of new classifier by the new class samples. The training loss function is defined as the common loss function for classification. That is:
\begin{equation}
\label{eqn:init}
\mathscr{L}(\mathbf{w}^N) = \left\{\sum\limits_{\mathbf{x} \in \mathbf{X}_{T}} L(f(\mathbf{x}, \mathbf{w}^N), y)\right\} + \lambda \cdot R(\mathbf{w}^N),
\end{equation}
where $L(\cdot, \cdot)$ is the prediction error and we use cross-entropy loss in our experiment. $R(\cdot)$ is a regularization term and we use \textit{L2}-norm in our experiment. For learning $\mathbf{w}^N$, we use the batched Stochastic Gradient Descent (SGD) and the $\mathbf{w}^N$ is initialized with the transferred classification parameters $\mathbf{w}^N_{trans}$.
\\

\noindent
\textbf{Tuning} \ We train the model classification parameters with new class samples, while adding a loss term to constrain the similarity of the transferred classification parameters and the final parameter:
\begin{equation}
\mathscr{L}(\mathbf{w}^N) = \left\{\sum\limits_{\mathbf{x} \in \mathbf{X}_{T}} L(f(\mathbf{x}, \mathbf{w}^N), y)\right\} + \lambda \cdot \norm{\mathbf{w}^N - \mathbf{w}^N_{trans}}{F}^2.
\end{equation}
Here, $\mathbf{w}_{trans}^N$ is the transferred parameter we obtain from the previous steps (\textit{i.e.} $\mathbf{w}_{new}^N$ in Equation \ref{eqn:final}). We still use the batched SGD method with a random initialization to solve for $\mathbf{w}^N$.
\\

\noindent
\textbf{Voting} \ This method is a weighted average for the transferred classification parameters and the learned model classification parameters. First, we learn a $\mathbf{w}^N_{model}$ using the Equation \ref{eqn:init} with random initialization. Then we get the final parameter by:
\begin{equation}
\mathbf{w}^N = \mathbf{w}^N_{trans} + \lambda \cdot \mathbf{w}^N_{model}.
\end{equation}
The hyper-parameter $\lambda$ serves as a voting weight.

\subsection{Complexity Analysis}
During the training process of our VAGER model, the main cost is to calculate the gradient of the loss function $\mathscr{L}(\mathbf{V}, \mathbf{T})$. For calculating the first derivative of $\mathscr{L}$ with respect to $\mathbf{V}$, the complexity per iteration is $O(nq \cdot max(p, n))$. As to the first derivative of $\mathscr{L}$ with respect to $\mathbf{T}$, the complexity per iteration is $O(nq \cdot max(p, q))$. While predicting the novel class, if we use Equation \ref{eqn:produce} for accelerating, we are able to pre-compute the $(\mathbf{V}^\top)^+$ for $O(nq^2)$ and for each novel class, the complexity of the predicting process is $O(q \cdot max(p, n))$.

\section{Experiments}
\subsection{Data and Experimental Settings}
In our experiments, we mainly use the ImageNet dataset \cite{ILSVRC15}, whose training set contains over 1.2 million images in 1,000 categories. We randomly divide the ImageNet training dataset into 800 base classes and 200 novel classes. 10 of the novel classes are used for validation to confirm the hyper-parameters and the other 190 novel classes are used for testing. We retrain the AlexNet on the 800 base classes as our base CNN model, where the training setting is the same as \cite{krizhevsky2012imagenet}. After training, we use the fc7 layer of AlexNet as the high-level representations for images and the parameters from fc7 to fc8 as the base classifiers' parameters (\textit{i.e.} matrix $\mathbf{W}$ in Equation \ref{eqn:base}) }. As our algorithm does not depend on the base model structure, we choose AlexNet as our base model in this paper. Moreover, when implementing our algorithm, we use 600 dimensions embedding space and the training hyper-parameter $\beta$ is set to 1.0.

We evaluate the performance of our algorithm from two aspects: Section 4.2 and Section 4.3 show a binary classification problem, where the new classifier is learned to classify the novel class (as positive samples) and all the base classes (as negative samples). This setting eliminates the relationship between novel classes and is convenient for us to validate each novel class independently, which is helpful to find the applicability of our algorithm, as Section 4.3 illustrates. In the training phase, we randomly select $k$ images as the training set for each novel class to simulate $k$-shot learning scenario. In the testing phase, given a novel class, we randomly select 500 images (no overlap with the training set) from it as the positive examples and randomly select 5 images from each base class of the ImageNet validation set as negative samples. To eliminate randomness, for any $k$-shot setting, we run 50 times and report the average result in the following experiments. Section 4.4 shows an $m$-way $k$-shot classification problem, where the new classifier is learned to classify among the $m$ novel classes, which is consistent with the classical setting in few-shot learning. In the training phase, we randomly select $m$ novel classes and select $k$ images from each of these classes as the training dataset. In the testing phase, we randomly select 5 images per novel class from the rest images as the testing dataset. The experiment will repeat 500 times under each $m$-way $k$-shot setting. 

The evaluating metric in our experiment is the Area Under Curve (AUC) of the Receiver Operating Characteristic (ROC) and the F1-score, which are widely used in binary classification. We report the average AUC and F1-score across all test classes. As to the $m$-way $k$-shot classification, we use average top-1 accuracy across $m$ novel classes.

We compare our method with the baselines below. The complete version of our method is VAGER+Voting.
\\
\noindent
\textbf{Logistic Regression (LR)} \hspace{0.1cm}
Common logistic regression model on novel classes. In the setting of multi-class classification, it becomes Softmax Regression. Note that LR is also equivalent to fine-tune the last layer of AlexNet.

\noindent
\textbf{Weighted Logistic Regression (Weighted-LR)} \hspace{0.1cm} Here we use the weighted average of the base classifiers' parameters as the classification parameters for a new class. The weights are calculated by an $L2$-normalized cosine similarities between the features of the novel class and 10 most similar base classes. This method can also be regarded as a visual analogy approach, but the transferring process is heuristic.

\noindent
\textbf{VAGER} \hspace{0.1cm} This is the VAGER algorithm without parameter refinement step.

\noindent
\textbf{VAGER(-Mapping)} \hspace{0.1cm} We directly learn the embedding by Equation \ref{eqn:base} without the first regression term. Then we use the above weighted-LR method in the embedding space instead of the original feature space. This method is used to evaluate the effectiveness of the mapping function.

\noindent
\textbf{VAGER(-Embedding)} \hspace{0.1cm} We directly train a regression model from the original feature space to the classification parameter space without the visual analogy graph embedding. This method is used to demonstrate the effectiveness of class node embedding over the visual analogy network.

Besides, we also consider some state-of-the-art algorithms as our baselines in multi-class classification setting, such as Model Regression Network (MRN)\cite{wang2016learning}, Matching Network (MatchingNet)\cite{vinyals2016matching}, Prototypical Network (ProtoNet)\cite{Snell2017Prototypical} and the method proposed in \cite{qiao2018few} (ActivationNet). Note that for MatchingNet and ProtoNet, we use a two-layer fully-connected neural network as the embedding architecture, which is consistent with \cite{WangCVPR2018a}.

\begin{table*}
  \renewcommand{\arraystretch}{1.0}
  \caption{Performance of different algorithms for $k$-shot binary classification problem}
  \label{tab:NewCategory2}
  \centering
  \begin{tabular}{ccccccccc}
  \toprule
\multirow{2}{*}{Algorithm}  &\multicolumn{2}{c}{\multirow{1}{*}{1-shot}} & \multicolumn{2}{c}{\multirow{1}{*}{5-shot}} & \multicolumn{2}{c}{\multirow{1}{*}{10-shot}} & \multicolumn{2}{c}{\multirow{1}{*}{20-shot}} \\
\cline{2-9}
& AUC & F1 & AUC & F1 & AUC & F1 & AUC & F1\\

    \midrule
    VAGER  & 0.8556 & 0.5292 & 0.9271 & 0.6491 & 0.9379 & 0.6721 & 0.9432 & 0.6850\\
    VAGER+Initializing  & 0.7662 & 0.3941 & 0.9030 & 0.6185 & 0.9338 & 0.6887 & 0.9461 & 0.7237\\
    VAGER+Tuning & 0.7923 & 0.4244 & 0.9098 & 0.6307 & 0.9365 & 0.7012 & 0.9466 & 0.7268\\
    VAGER+Voting & \textbf{0.8718} & \textbf{0.5671} & \textbf{0.9425} & \textbf{0.7039} & \textbf{0.9543} & \textbf{0.7343} & \textbf{0.9607} & \textbf{0.7510}\\
    VAGER(-Mapping)  & 0.8261 & 0.4551 & 0.8526 & 0.4807 & 0.8726 & 0.5179 & 0.8897 & 0.5394\\
    VAGER(-Embedding)  & 0.7922 & 0.4335 & 0.9032 & 0.6015 & 0.9183 & 0.6347 & 0.9393 & 0.6788\\
    \midrule
    LR &  0.7705 & 0.3994 & 0.8885 & 0.5882 & 0.9134 & 0.6421 & 0.9341 & 0.6877\\
    Weighted-LR & 0.8440 & 0.4775 & 0.8458 & 0.4813 & 0.8509 & 0.4835& 0.8468 & 0.4801\\ 
    MRN  & 0.8083 & 0.4511 & 0.9175 & 0.6653 & 0.9361 & 0.7133 & 0.9474 & 0.7388\\
  \bottomrule
\end{tabular}
\footnotetext[1]{* Our Algorithm}
\end{table*}

\subsection{Binary Classification}

In this section, we evaluate how well the classifiers learned by our method and other baselines can perform in novel classes on binary classification setting. 

The results are shown in Table \ref{tab:NewCategory2}. In all low-shot settings, our method VAGER+Voting consistently performs the best in both AUC and F1 metrics. In contrast, LR performs the worst in 1-shot setting, which demonstrates the importance of generalization from base classes when the new class has very few samples. MRN does not work well in most settings, demonstrating that its basic hypothesis that the classification parameters trained by large samples and small samples respectively are correlated does not necessarily hold in real data. By comparing VAGER+Voting with the other five variant versions of our method, we can safely draw the conclusion that the major ingredients in our method, including network embedding for low dimensional representations, mapping function for transforming embedding space to classification parameter space, as well as the refinement strategy are necessary and effective and the results support that
the Voting strategy performs the best in our scenario.

\begin{figure}[H]
\centering
\includegraphics[width=3.0in]{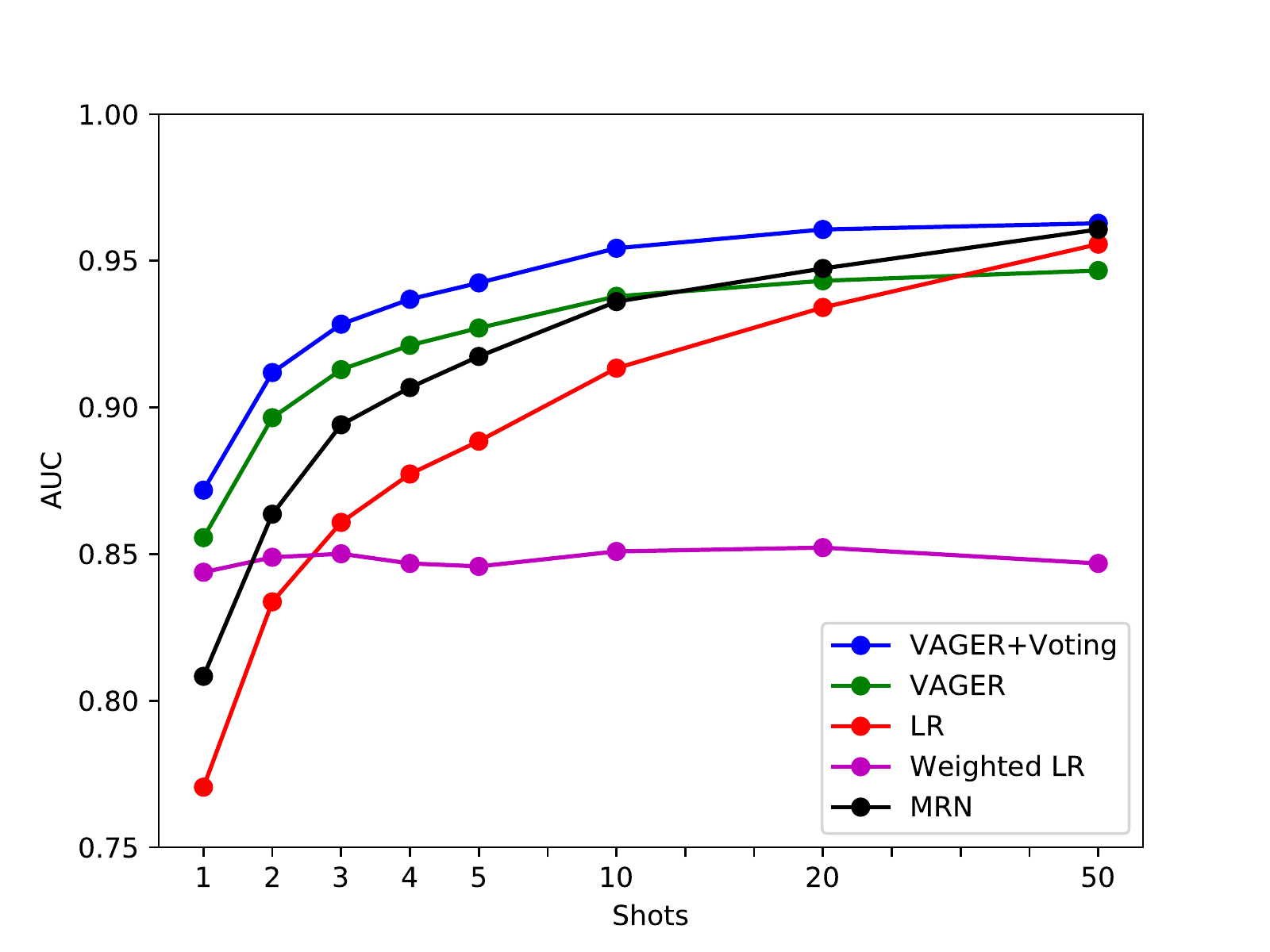}
\caption{The change of performance as the number of shots increases in binary classification.}
\label{Fig:Shot}
\end{figure}

Furthermore, we compare the performances of these methods in different low-shot settings, and the results are shown in Figure \ref{Fig:Shot}. Our method consistently performs the best in all settings, and the advantage of our method is more obvious when the novel classes have less training samples. Especially, by comparing our method and LR, we can see that LR needs about 20 shots to reach AUC 0.9, while we only need 2 shots, indicating that we can save $90\%$ training data. An interesting phenomenon is that the performance of Weighted-LR does not change as the shot number increases. The main reason is that the heuristic rule is not flexible enough to incorporate new information, which demonstrates the importance of learning to learn, rather than rule-based learning.

\subsection{Insightful Analysis}
Although our method performs the best in different settings, the failure cases are easy to find. We are interested in the following questions: (1) What are the typical failure cases? (2) What is the driving factor that controls the success of generalization? (3) Whether the generalization process is explainable?

In order to answer the above questions, we further conduct an insightful analysis. We randomly select 10 novel classes, and list the performance of our method compared with LR in one-shot setting on these classes, as shown in Table \ref{tab:MultiClassOneshot}. It's obvious that the effect of generalization is notable in 9 of them, but in the bubble class, the generalization plays a negative role. 

To discover the driving factor controlling success or failure of the generalization, we define and calculate the similarity ratio (SR) of a novel class with the base classes by:
\begin{equation}
SR = \frac{Average\ Top\text{-}K\ Similarity \ with\ Base\ Classes}{Average\ Similarity\ with\ Base\ Classes}
\end{equation}
Here the similarity of two classes is calculated by Equation \ref{equ:sim}. Intuitively, if a new class is similar with the top-$K$ base classes, while dissimilar with the remained base classes, its Similarity Ratio will be high, meaning that this new class can benefit more from the base classes.

\begin{figure}
\centering
\includegraphics[width=3.0in]{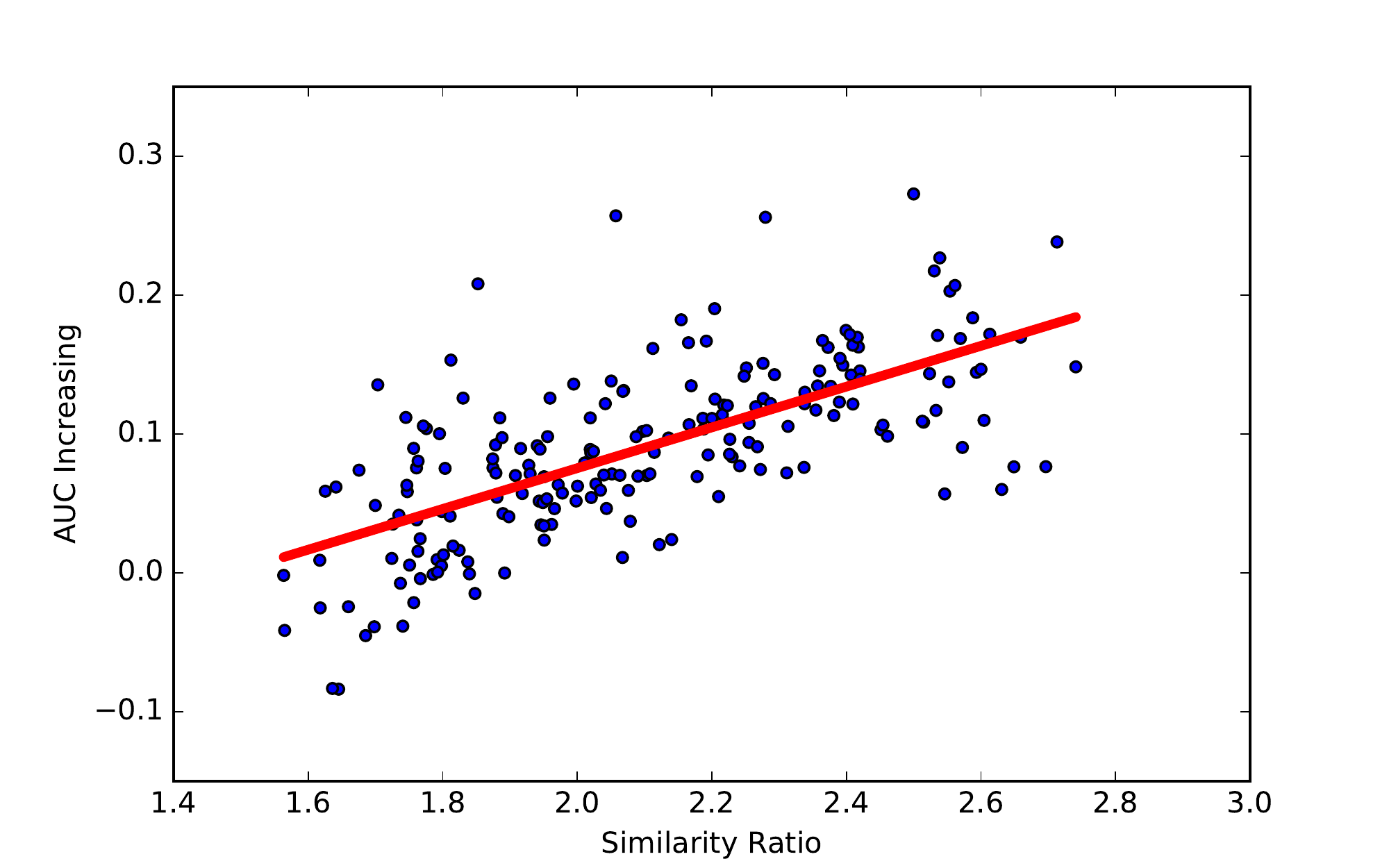}
\caption{Linear regression of AUC improvement on Similarity Ratio for all novel classes}
\label{Fig:Regression}
\end{figure}

For each new class, we calculate the relative improvement in AUC of our method over non-transfer method LR in 1-shot setting, and do linear regression over its Similarity Ratio with $K=10$. The dependent variable indicates the success degree of generalization. And we set $K = 10$. We plot the similarity ratio and relative improvement of all novel classes in Figure \ref{Fig:Regression}. We can see that the relative improvement in a new class is positively correlated with the similarity ratio of the new class, with $95\%$ confidence interval for the correlation coefficient range between $0.124$ and $0.169$ and $R^2 = 0.45$, showing that the SR ratio could explain 45\% of the dependent variable. 

The results fully demonstrate that our method is consistent with the notion of human-like learning: First, we can learn a new concept faster if it is more similar to some previously learned concepts. (\textit{i.e.} Leading to the increase of the numerator of the Similarity Ratio). Second, we can learn a new concept faster if we have learned more diversified concepts (\textit{i.e.} Leading to the decrease of the denominator of the Similarity Ratio). This principle can also be used to guide the generalization process and help to determine whether a new class is fit for generalization.

\begin{table}[!t]
  \renewcommand{\arraystretch}{1.0}
  \caption{Comparison of VAGER and LR over novel classes with 1-shot binary classification setting}
  \label{tab:MultiClassOneshot}
  \centering
  \begin{tabular}{ccc}
    \toprule
    Category&LR (No Transfer)&VAGER (Transfer)\\
    \midrule
    Jeep & 0.8034 & \textbf{0.9469}\\
    Zebra & 0.8472 & \textbf{0.9393}\\
    Hen & 0.7763 & \textbf{0.8398}\\
    Lemon & 0.6854 & \textbf{0.9583}\\
    Bubble & \textbf{0.7455} & 0.7041\\
    Pineapple & 0.7364 & \textbf{0.8623}\\
    Lion & 0.8305 & \textbf{0.9372}\\
    Screen & 0.7801 & \textbf{0.9056}\\
    Drum & 0.6510 & \textbf{0.6995}\\
    Restaurant & 0.7806 & \textbf{0.8787}\\
  \bottomrule
\end{tabular}

\footnotetext[1]{* Our Algorithm}
\end{table}

\begin{figure}
\centering
\includegraphics[width=3.0in]{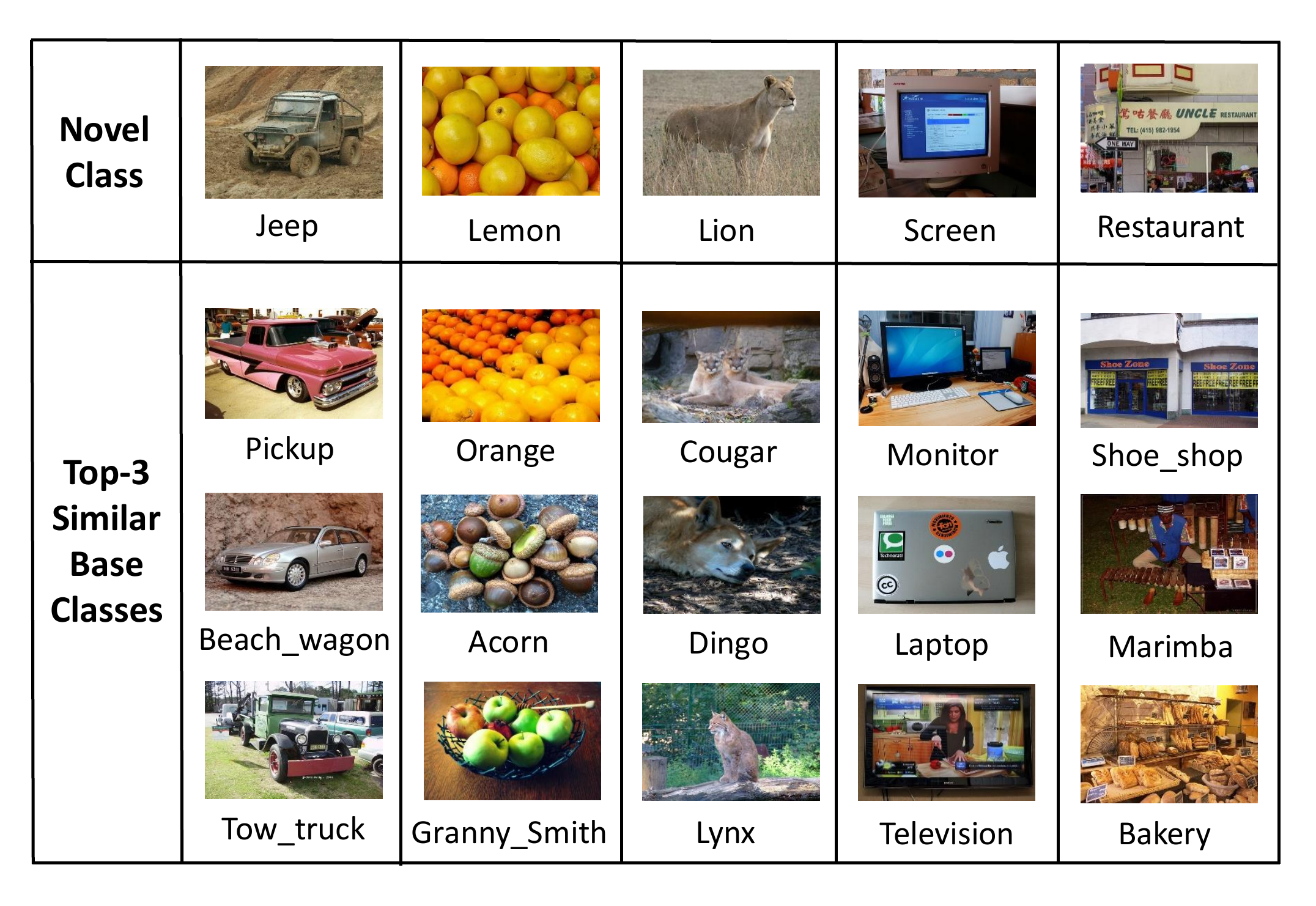}
\caption{Top-3 most similar base classes to novel class on embedding layer in 5-shot setting.}
\label{Fig:Result}
\end{figure}

\begin{table*}
  \renewcommand{\arraystretch}{1.0}
  \caption{Top-1 Accuracy for m classes 1-shot problem}
  \label{tab:multigroup}
  \centering
  \begin{tabular}{ccccccccc}
  \toprule
Algorithm & 10\,cls/G1 & 10\,cls/G2 & 10\,cls/G3 & 10\,cls/G4 & 10\,cls/G5 & 30\,cls/G1 & 50\,cls/G1 & 100\,cls/G1\\

    \midrule
    VAGER+Voting & \textbf{67.59\%} & \textbf{63.96\%} & \textbf{58.02\%} & \textbf{51.27\%} & \textbf{56.24\%} & \textbf{40.73\%} & \textbf{38.69\%} & \textbf{28.38\%}\\
    LR & 61.97\% & 59.72\% & 52.97\% & 47.51\% & 52.01\% & 37.32\% & 34.75\% & 23.94\%\\
    Weighted-LR & 63.13\% & 60.09\% &50.32\% & 46.13\% & 49.81\% & 36.77\% & 34.64\% & 23.60\%\\
    MRN & 64.55\% & 61.82\% & 54.74\% & 48.85\% & 54.54\% & 39.43\% & 37.78\% & 27.16\%\\
    MatchingNet & 65.69\% & 61.74\% & 57.13\% & 48.56\% & 54.34\% & 39.04\% & 37.05\% & 27.21\%\\
    ProtoNet & 47.98\% & 47.18\% & 40.20\% & 35.86\% & 41.55\% & 30.15\% & 28.12\% & 21.28\%\\
    ActivationNet & 65.04\% & 62.42\% & 55.62\% & 48.61\% & 53.85\% & 40.15\% & 37.41\% & 27.68\%\\
  \bottomrule
\end{tabular}
\end{table*}
Finally, we validate whether the generalization process is explainable. Here we randomly select 5 novel classes, and for each novel class, we visualize the top-3 base classes that are most similar with the novel class in the visual analogy graph, as shown in Figure \ref{Fig:Result}. In our method, these base classes have a large impact on the formation of the new classifier. We can see that the top-3 base classes are visually correlated with the novel classes, and the generalization process can be very intuitive and explainable.
\begin{figure}
\centering
\includegraphics[width=3.0in]{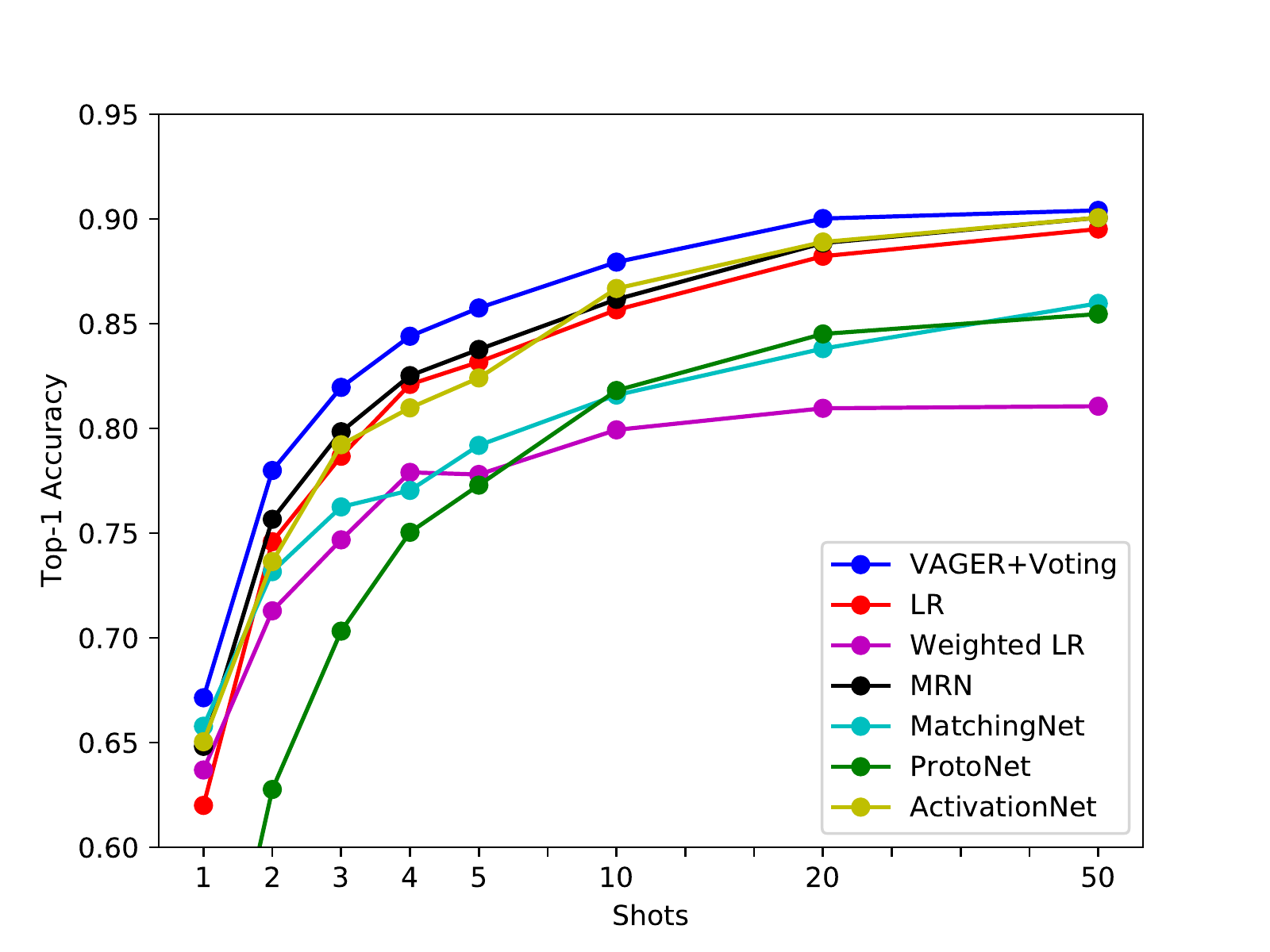}
\caption{Change of performance as shot number increases in 10 classes 1-shot multi-class classification problem.}
\label{Fig:Shot2}
\end{figure}
\subsection{Multi-class Classification}
{In this section, we mainly show the performance of the experiments on multi-class classification. We will show that our algorithm performs well from three aspects. All baselines in Section 4.2 are extended to multi-class classification version in these experiments.

The first experiment is to validate the robustness of our algorithm. We randomly select 10 categories from the novel test categories and learn to distinguish these 10 categories on the 1-shot setting. We repeat random selections five times and the result is shown on the first 5 columns in Table \ref{tab:multigroup}. Our VAGER+Voting performs the best in all 5 groups, with promotion of around $2\%$ of average top-1 accuracy, which demonstrates that our method is robust whatever the novel classes are.

The second experiment is to evaluate our method on different numbers of novel classes. We design an 10/30/50/100-way 1-shot setting. The result is shown in the last four columns in Table \ref{tab:multigroup}. As the result shows, our algorithm consistently gets the best performance.

The third experiment is to evaluate our method on different shots. We control the number of novel classes and change the number of shots used for learning novel classifiers. We randomly choose 10 novel classes and test the performance of our algorithm and baselines on 1/2/3/4/5/10/20/50 shots. The result is shown in Figure \ref{Fig:Shot2}. In all scenarios, our algorithm performs the best. Although MatchingNet and ProtoNet could do better on small dataset like Omniglot \cite{Lake2013One} and MiniImageNet\cite{vinyals2016matching}, in large-scale dataset, however, their performances are not satisfactory. One plausible reason is that the effectiveness of their meta-learning mechanism is limited when the embedding architecture is representative enough. On the other hand, MRN and ActivationNet adopt learning to learn mechanism as well. The advantage of our method over these two baselines is attributed to learning by analogy mechanism that is inspired by human learning.

\section{Conclusions}
In this paper, we investigate the problem of learning to learn image classifiers and explore a new human-like learning mechanism which fully leverages the previously learned concepts to assist new concept learning. In particular, we organically combine the ideas of learning to learn and learning by analogy and propose a novel VAGER model to fulfill the generalization process from base classes to novel classes. From the extensive experiments, it shows that the proposed method complies with human-like learning and provides an insightful and intuitive generalization process. 

\noindent \textbf{Acknowledgement:} This work was supported in part by National Program on Key Basic Research Project (No. 2015CB352300), National Natural Science Foundation of China (No. 61772304, No. 61521002, No. 61531006, No. 61702296), National Natural Science Foundation of China Major Project (No.U1611461), the research fund of Tsinghua-Tencent Joint Laboratory for Internet Innovation Technology, and the Young Elite Scientist Sponsorship Program by CAST.

{\small
\bibliographystyle{ieee_fullname}
\bibliography{sigproc}
}

\end{document}